# Hybrid Artificial Intelligence Methods for Predicting Air Demand in Dam Bottom Outlet

Aliakbar Narimani [1], Mahdi Moghimi [1] and Amir Mosavi[2,*]

[1] School of Mechanical Engineering, Iran University of Science and Technology, Tehran, Iran; a_narimanizamanabadi@yahoo.com; moghimi@iust.ac.ir

[2] School of Economics and Business, Norwegian University of Life Sciences, 1430 Ås, Norway;

**Abstract:** In large infrastructures such as dams, which have a relatively high economic value, ensuring the proper operation of the associated hydraulic facilities in different operating conditions is of utmost importance. To ensure the correct and successful operation of the dam's hydraulic equipment and prevent possible damages, including gates and downstream tunnel, to build laboratory models and perform some tests are essential (the advancement of the smart sensors based on artificial intelligence is essential). One of the causes of damage to dam bottom outlets is cavitation in downstream and between the gates, which can impact on dam facilities, and air aeration can be a solution to improve it. In the present study, six dams in different provinces in Iran has been chosen to evaluate the air entrainment in the downstream tunnel experimentally. Three artificial neural networks (ANN) based machine learning (ML) algorithms are used to model and predict the air aeration in the bottom outlet. The proposed models are trained with genetic algorithms (GA), particle swarm optimization (PSO), i.e., ANN-GA, ANN-PSO, and ANFIS-PSO. Two hydrodynamic variables, namely volume rate and opening percentage of the gate, are used as inputs into all bottom outlet models. The results showed that the most optimal model is ANFIS-PSO to predict the dependent value compared with ANN-GA and ANN-PSO. The importance of the volume rate and opening percentage of the dams' gate parameters is more effective for suitable air aeration.

1. Introduction

The bottom outlet and spillways have a significant role in the dam structure that have been used for several applications, such as sediment flushing and emptying the reservoir [1-4]. Although scientists have carried out many research types, there is not enough certainty about proper design for preventing choked flow in downstream and cavitation on the surface and wall [2]. Kalinske [5] carried out the first experiments about air entrainment by calculating air ventilation at hydraulic jumps in a circular conduit. They presented an equation for air demand as Equation 1:

$$\beta = \frac{Q_{air}}{Q_{water}} = 0.0066(Fr - 1)^{1.4} \qquad (1)$$

Another method implemented by Sharma [6] is to classify the flow aeration after the gate to three groups of a hydraulic jump, spraying, and free-surface flow. After a decade, Falvey carried out a thorough research on cavitation in spillways and chutes and devised some air ventilation methods from the atmosphere in [7]. Their results suggested Wu et al [8] designed a method to place an aerator in a high head dam showed that the conduit may endure cavitation if the dam head is higher than 355 m at the Longtan Hydropower Station. Another study designed by Wu et al. [8] carried out the partial opening gate's hydraulic features in a bottom outlet. They found that the discharge coefficient at a smaller gate opening is the same as when the gate is completely open. To determine the effect of size-scale on air entrainment, Mortensen et al. [9] measured air demand in four different-sized pipes at the same Froude number. The results showed that the air entrainment percentage was approximately similar in different conduit sizes when the hydraulic jump completely happened in conduit. Besides physical investigations, a numerical study was carried out to evaluate air demand in bottom outlets and simulate air distribution in gated tunnels [10]. Another numerical study on the dam bottom outlet was performed to demonstrate that there is no cavitation zone in the Shahryar dam in Iran [11]. To find the highest air demand, researchers investigated the impact of gate opening on air velocity in Darian's dam's bottom outlet. They demonstrated that air ventilation peaked in 80% service gate opening [12]. A study performed a series of experiments in 2014 to show that air discharge peaked in 60% emergency gate opening. They measured the aeration ratio and cavitation index in their study as well [13].

A wide study was carried out on the Berg river dam model to investigate the air demand in the bottom conduit. The outcome shows that the air velocity was independent of the time of valve closure [14]. Among all hydraulic parameters, flow rate effect on air demand was investigated in the spillway of highest arched dam in the world to observe that increasing flow rate can lead to rising air demand [15]. Recently, Zhang et al. [16] investigated a novel model on air concentration and pressure along the bottom conduit. The model divided the downstream conduit into four zones: the cavity zone, the impact zone, the equilibrium zone, and the far zone. They also presented the amount of air entrained in each zone. Although the previous experimental and numerical studies show their significance on the hydraulic performance in the dam bottom outlet, it lacks cost efficiency and accuracy of detection.

On the other hand, natural aeration will happen in spillways that can decrease the presence of cavitation [17]. Rebollo et al. experimentally investigated how aeration influence the energy dissipation. Their results showed that higher aeration can cause less energy loss in spillways [18]. A numerical investigation was carried out to show the effect of physical parameters in stepped spillways on aeration influence. The results showed that the height of the first step can be a measure to prevent cavitation erosions in steps [19]. Another numerical approach in spillways was presented by Badas et al. which demonstrated that scale effects can cause the error in the aeration rate of the spillways jet [20]. Yang et al. carried out an experimental and numerical research on spillways and bottom outlets. They showed flow release from just bottom outlets is the most critical situation in the case study and air entrainment plays an effective role in reducing cavitation [21]. Aydin also investigated efficiency of the aeration in bottom of spillways in 2018 and presented a formula to estimate the air entrainment rate in the aerators of the bottom-inlet [22]. Hien presented an air entrainment model in stepped spillways to effectively simulate the aeration zone. This model was also able to simulate flow over a complex geometry [23].

To increase the accuracy of detection and reach low-cost efficiency, computational intelligence [24, 25] approaches apply to predict the flow behavior in dam bottom outlet. Machine learning (ML) methods adapt to deal with this complex problem. Recently, several studies have employed ML techniques for the prediction of aerator air demand. Zounemat-Kermani et al (2013) proposed the ANFIS model to handle the air demand in a low-level outlet. The back-propagation and least square estimation have been combined to generate a hybrid learning algorithm for handling the linear and non-linear parameters by the ANFIS. The developed ANFIS model was compared with feed-forward multi-layered perceptron and multiple linear regression in terms of root mean square error (RMSE), Nash–Sutcliffe efficiency, and correlation coefficient as the performance factors. According to the results, the developed ANFIS model could successfully eliminate the prediction error by about 36 and 74 % compared with feed-forward multi-layered perceptron and multiple linear regression, respectively [26]. Wu and Fan (2013) developed a radial basis function (RBF) neural network to estimate the bottom slopes in different conditions of flow and geometry from different discharge tunnels with safe operation. Results were evaluated by calculating the error values according to the parameters related to the slopes. Findings claimed that the proposed RBF neural network could successfully cope with the modeling task [27]. Najafi et al (2012) developed ANFIS, Fuzzy, and the genetic fuzzy system to estimate the air demand in the gated tunnels to prevent cavitation attack. A comparison for the selection of the best model was performed by RMSE and correlation coefficient values. According to the results, the ANFIs model could provide a higher prediction performance compared with others [28]. Taherei Ghazvinei et al (2017) employed a hybrid support vector machine integrated by discrete wavelet transform to evaluate an intake structure and its safety by the prediction of the head loss at the inlet and outlet section of the horizontal intake structure. Results of the developed method were compared with ANN and genetic programming in terms of RMSE, correlation coefficient, and determination coefficient. According to the results, in average, the developed hybrid method (support vector machine integrated by discrete wavelet transform) improved the correlation coefficient about 4 %, the RMSE about 13 %, and determination coefficient about 2 % in comparison with ANN and genetic programming [29].

The modeling of air demand in the bottom outlet of dams can be considered as a complex problem that requires efficient non-linear meta-heuristic methods to be addressed [30]. According to the studies developed for the prediction of air aeration parameters, the single ANN and ANFIS models have been frequently employed by the studies.

The present study aims to develop a hybrid ANN and ANFIS models by combining GA and PSO optimizers for the prediction of the air entrainment in the dam bottom outlet. The proposed ensemble ML algorithms can increase the accuracy of prediction and reduce the false alarm rate. Thus, we employed hybrid ANN-GA, ANN-PSO, and ANFIS-PSO to estimate the air demand in the bottom outlet of dams. The main contributions of this paper can be summarized as follows:
- To experimentally investigate air demand in bottom outlet and collect related data
- To develop and propose the hybrid models according to the generated dataset
- To evaluate the training and testing phases for the developed models

The current study is organized into six sections. In Section 2, experimental setup has been presented. Section 3 is dedicated to the proposed methods whereas Section 4 presents the model performance evaluation. Results and discussion of this paper are provided in Sections 5 and Finally, the research conclusion is presented in Section 6.

**2. Materials and methods**

Air entrainment in the model of six dams was studied experimentally in the applied hydrodynamic laboratory of Iran university of science and technology (IUST). The bottom outlets with aerator have been used with the minimum range of gate opening (10%) in our case study which is 6 dams. The location and specification of the dam are given in Figure 1 and Table 1, respectively. Figure 1 shows the locations of dam 1 (Safarood dam), dam 2 (Balarood dam), dam 3 (Sardasht dam), dam 4 (Silve dam), dam 5 (Talvar dam) and dam 6 (Kucheri dam).

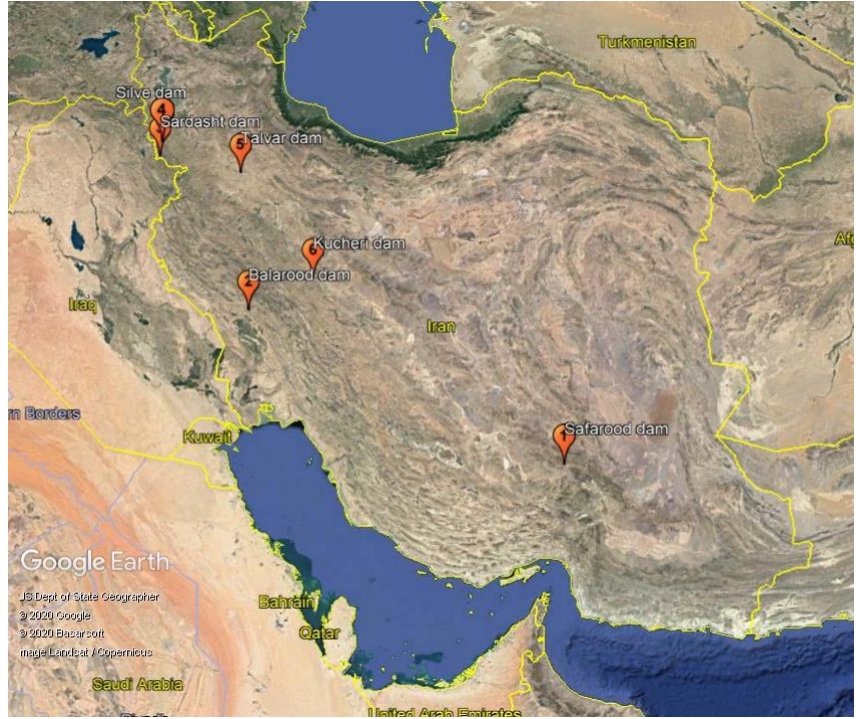

**Figure 1.** Approximate location of understudied dams in Iran [31]

Table 1 illustrates flow rate and head of dams in prototype and besides indicates physical characteristics of dam such as downstream length and gate parameters used in operation conditions.

**Table 1.** specification of dams

| | | Flow (m³s⁻¹) | | | | Water head [He] (m) | | | Downstream length [L] (m) | Gate parameters | |
|---|---|---|---|---|---|---|---|---|---|---|---|
| | | Water | | Air | | | | | | Range of opening [O] (%) | Dimensions [H×W] (m×m) |
| No. | Dam Name | min | max | min | max | min | normal | max | | | |
| 1 | Safarood | 8.7 | 48.2 | 6.8 | 18.9 | - | - | 59.4 | 12-60 | 20-100 | 1.47×1.19 |
| 2 | Balarood | 2.2 | 44.8 | 5.2 | 18.0 | - | 58.0 | 69.0 | 40 | 10-100 | 1.39×1.17 |
| 3 | Sardasht | 14.1 | 225.0 | 15.7 | 54.0 | 42.3 | 87.2 | 95.2 | 60 | 10-100 | 2.80×2.23 |
| 4 | Silve | 4.6 | 96.3 | 40.0 | 210.3 | 27.6 | - | 56.4 | 40 | 10-100 | 2.00×1.89 |
| 5 | Talvar | 15.1 | 179.4 | 88.0 | 152.7 | - | 56.5 | - | 60 | 10-100 | 3.12×2.14 |
| 6 | Kucheri | 27.7 | 243.2 | 27.2 | 71.7 | - | 64.0 | - | 30 | 10-100 | 2.79×2.29 |

Even though there is a great demand for using high-performance computers to solve turbulent flow, not so precise answers have been provided yet. Consequently, verification and comparison with empirical data are required. Accordingly, empirical experiments are the best approaches for analyzing phenomena in fluid mechanics. However, many phenomena need to be tested to achieve their goal, which can be costly. Dimensional analysis in addition to the ability to set up laboratory models, alongside data analysis and interpretation, provides a simple technique to reduce the number of required trials [32].

As it has been indicated in Figure 2, effective parameters in a dam are presented. Moghimi et al. [12] derived all dimensionless numbers based on the Buckingham theorem and showed that effective dimensionless numbers are $\frac{Q}{A\sqrt{2gH}}$ and $\frac{A}{A_0}$. They presented a formula as Equation 2.

$$\frac{Q}{A\sqrt{2gH}} = \Phi\left(\frac{A}{A_0}\right) \qquad (2)$$

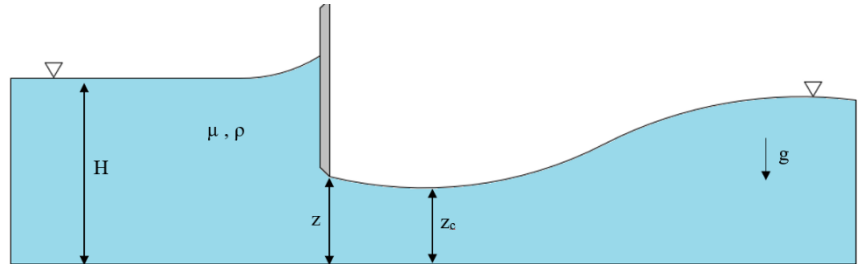

**Figure 2.** dam schematic and effective parameters

Figure 2 indicates dam parameters such as dam head (H), dynamic viscosity of water (μ), water density (ρ), height of gate section (z), height of minimum section (zc) and gravitational acceleration (g).

Experiments have been carried out in the closed-loop in the applied hydrodynamic laboratory of IUST. Figure 3 indicates an overview of the circuit.

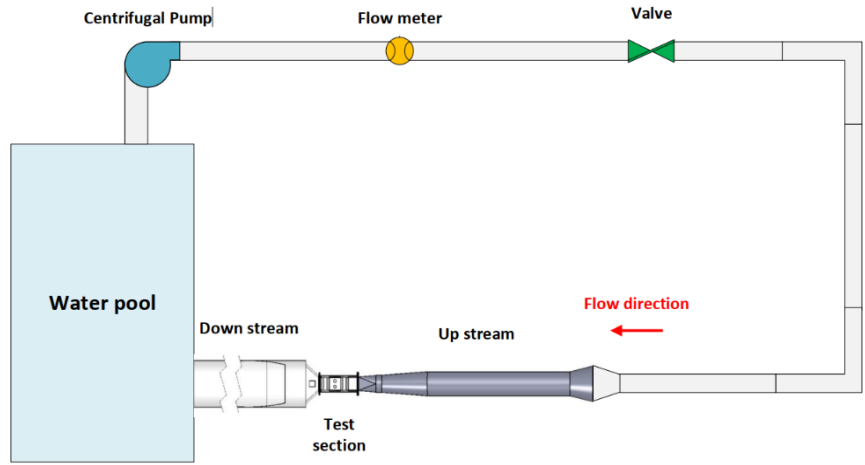

**Figure 3.** Closed-loop for dam model tests

To measure the instantaneous and mean pressure values, 64 piezometers were used in the dam bottom outlet model. These piezometers are installed in different locations of the conduit as required. All of these piezometers are connected by tubes with a diameter of 4 mm outside the pressure box embedded near closed-loop, measured and recorded gauge, and absolute pressure, from various locations in the loop. To measure flow rate and water velocity during the test, a Proline Promag 10 P electromagnetic flow meter was used from Endress Hauser Company. Its information was recorded simultaneously with other parameters by a computer. After the centrifugal pump, a globe valve was installed to vary the flow rate to make different conditions during the experiments. To record, store and display information, a program called Hydrolab logger was written in the applied hydrodynamic laboratory of IUST. This software has two calibration constants that are different for various pressure gauges that need to be adjusted. The program can also set the data rate and data time. A hot wire sensor (model AM-4204) has been employed to record the air velocity between gates and after the service gate.

*2.1. Proposed method*

In the present study, the modeling process was performed using three robust hybrid methods including ANN-GA, ANN-PSO, and ANFIS-PSO. ANN combined with GA and PSO algorithms to train the parameters of ANN by different numbers of neurons in the hidden layer (from 8 to 16 with interval 4 neurons) in the presence of the population sizes for GA and PSO (50, 100, and 150). Figure 4 (regenerated from [33]) presents the architecture of the ANN method. In the training step, 70 percent of total data were employed, and the rest of the data were employed for the testing data.

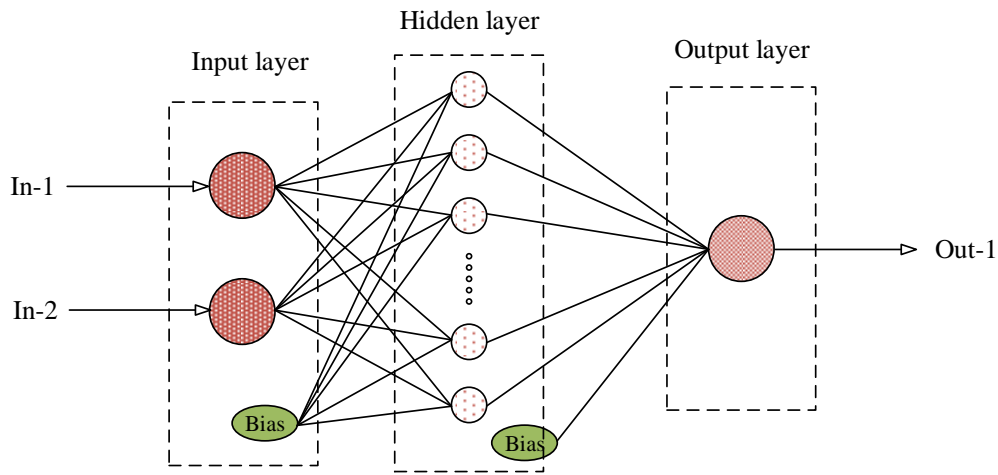

**Figure 4.** the architecture of the ANN method

ANFIS parameters are formed as variables of the PSO and the mean-squared-error (MSE) is utilized as the performance factor (fitness factor) of PSO. Training ANFIS was performed to adapt the ANFIS parameters such that the fitness function, according to the MSE of the ANFIS, to be reduced by the PSO algorithm. Membership function (MF) type and the population size of the PSO method have been considered as the parameters which affect the performance of predictions. Three types of MFs including triangular, G. Bell, and Gaussian types were selected by population size 50, 100, and 150. To train ANFIS-PSO, 70% of data selected and it utilized 49 rules with linear output MF type. Figure 5 presents the architecture of the ANFIS method (regenerated from [33]).

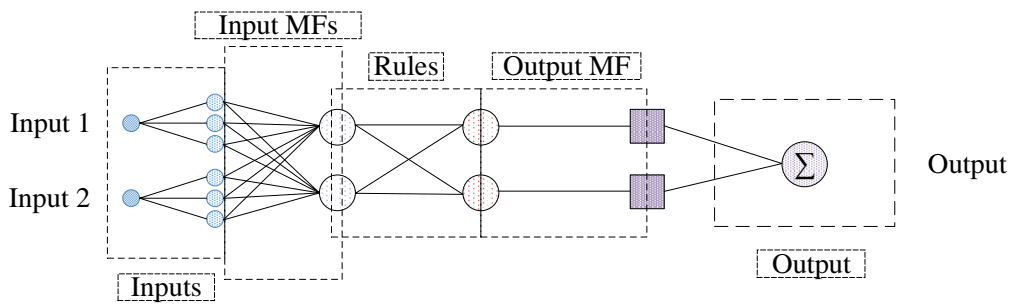

**Figure 5.** the architecture of the ANFIS method

What can be seen in Figure 5 (Regenerated from [34]) is two inputs variables such as volume rate and opening percentage of the gate and air velocity as an output.

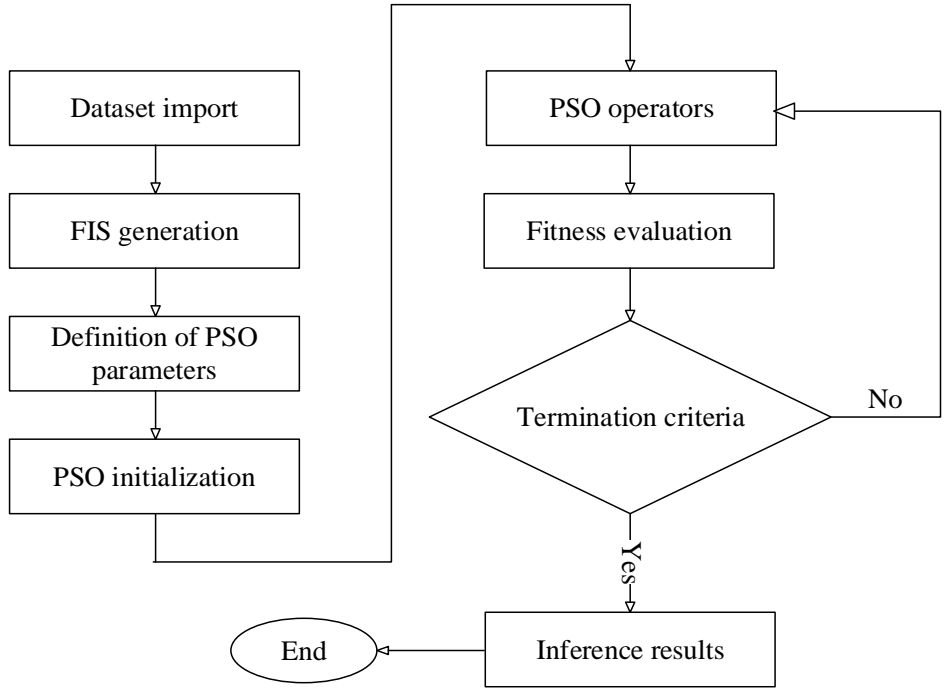

**Figure 6**. the algorithm of hybrid ANFIS-PSO method

Figure 6 presents the main flowchart of the optimization of a training process using ANFIS PSO.

*2.2 Model performance evaluation*

Evaluation metrics can present an index for comparing the performance and accuracies of the predicting models. These models compare the output of models with the target values and calculate the differences [35, 36]. (and eventually) the lower differences refer to the highest performance. In the present study, four frequently used performance factors including root mean square error (RMSE), mean square error (MSE), correlation coefficient (CC) and sustainability index (SI) have been employed for the best judgment around finding the model with a high performance (Eq. 3 to 6).

$$RMSE = \sqrt{\frac{1}{n}\sum_{i=1}^{n}(O_i - P_i)^2} \qquad (3)$$

$$MSE = \frac{1}{n}\sum_{i=1}^{n}(O_i - P_i)^2 \qquad (4)$$

$$CC = \frac{\left(\sum_{i=1}^{n} O_i P_i - \frac{1}{n}\sum_{i=1}^{n} O_i \sum_{i=1}^{n} P_i\right)}{\left(\sum_{i=1}^{n} O_i^2 - \frac{1}{n}(\sum_{i=1}^{n} O_i)^2\right)\left(\sum_{i=1}^{n} P_i^2 - \frac{1}{n}(\sum_{i=1}^{n} P_i)^2\right)} \qquad (5)$$

$$SI = \frac{\sqrt{\frac{1}{n}\sum_{i=1}^{n}(P_i - O_i)^2}}{\overline{O}} \qquad (6)$$

Where O refers to the target values, P refers to the predicted values and n refers to the number of data.

**3. Results**

*3.1. Experimental results*

To study air demand in dams, three parameters should be changed namely head, volume rate and opening percentage that amount of air entrainment would be evaluated. Experimental tests have been performed for all dams in operating conditions with suitable scale and reiteration of the tests has been checked by performing several tests to obtain real results. Figure 7 shows an overview of the experiments and model in Kucheri and Talvar dam.

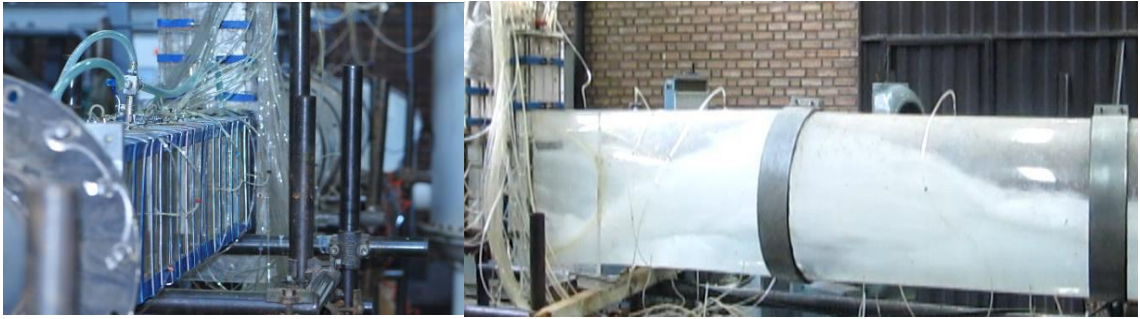

**Figure 7.** Talvar dam model (left) and Kucheri's downstram in maximum head

*3.2. Simulation results*

This section presents the training results. Training is developed by different types of ANN-GA, ANN-PSO, and ANFIS-PSO to choose the best architecture to be employed in testing phase. Table 1 presents the training results in terms of RMSE, CC and SI as the performance factors for the developed ANN-GA and ANN-PSO models.

**Table 2.** results of training phase for ANN-GA and ANN-PSO

| Model name | Number of neurons | Pop. Size | RMSE | CC | SI |
|---|---|---|---|---|---|
| ANN-GA | 8 | 50 | 5.958 | 0.581 | 0.495 |
| | *12* | *50* | *5.835* | *0.604* | *0.490* |
| | 16 | 50 | 5.938 | 0.609 | 0.522 |
| | 8 | 100 | 6.146 | 0.554 | 0.480 |
| | 12 | 100 | 7.595 | 0.466 | 0.593 |
| | 16 | 100 | 6.250 | 0.524 | 0.525 |
| | 8 | 150 | 6.459 | 0.485 | 0.555 |
| | 12 | 150 | 5.898 | 0.599 | 0.486 |
| | 16 | 150 | 6.333 | 0.511 | 0.522 |
| ANN-PSO | *8* | *50* | *5.285* | *0.693* | *0.440* |
| | 12 | 50 | 5.545 | 0.656 | 0.443 |
| | 16 | 50 | 6.025 | 0.576 | 0.471 |
| | 8 | 100 | 5.494 | 0.669 | 0.436 |
| | 12 | 100 | 5.377 | 0.690 | 0.455 |
| | 16 | 100 | 6.282 | 0.530 | 0.563 |

| | 8 | 150 | 6.018 | 0.570 | 0.502 |
|---|---|---|---|---|---|
| | 12 | 150 | 5.459 | 0.670 | 0.462 |
| | 16 | 150 | 5.894 | 0.604 | 0.490 |

Table 2 shows the ANN-GA with a number of neurons 12 and population size 50 provided lower RMSE (5.834) and SI (0.49) and higher CC (0.604) in comparison with other architectures of ANN-GA and ANN-PSO with the number of neurons 8 and population size 50 provided lower RMSE (5.384) and SI (0.43) and higher CC (0.692) compared with other architectures of ANN-PSO. In training phase, ANN-PSO could successfully provide higher performance over the ANN-GA model. Table 3 presents the training results for ANFIS-PSO. The results show that the higher performance is related to the ANFIS-PSO with MF type triangular at population size 100 with RMSE 3.227, CC 0.898 and SI 0.27 over other architectures of ANFIS-PSO models.

Table 3. results of training phase for ANFIS-PSO

| Model name | MF type | Pop. Size | RMSE | CC | SI |
|---|---|---|---|---|---|
| ANFIS-PSO | Triangular | 50 | 3.265 | 0.896 | 0.272 |
| | *Triangular* | **100** | 3.227 | 0.898 | 0.270 |
| | Triangular | 150 | 3.245 | 0.897 | 0.272 |
| | G. bell | 50 | 5.571 | 0.649 | 0.466 |
| | G. bell | 100 | 5.535 | 0.665 | 0.461 |
| | G. bell | 150 | 5.557 | 0.651 | 0.463 |
| | Gaussian | 50 | 3.759 | 0.858 | 0.312 |
| | Gaussian | 100 | 3.270 | 0.896 | 0.274 |
| | Gaussian | 150 | 3.604 | 0.871 | 0.300 |

The selected models from the training phase were employed in the testing phase. Table 4 presents the results related to the testing phase by comparing the performance of ANN-GA, ANN-PSO, and ANFIS-PSO models.

Table 4. results for the testing phase

| Model name | RMSE | CC | SI |
|---|---|---|---|
| ANN-GA | 5.208 | 0.586 | 0.415 |
| ANN-PSO | 4.977 | 0.595 | 0.440 |
| ANFIS-PSO | 3.097 | 0.756 | 0.252 |

Figure 8 presents the plot diagrams for a 1:1 comparing of the output and target values in the testing phase. As is clear, ANFIS-PSO with the highest determination coefficient 0.75 could successfully increase the performance of the prediction about 52% and 54% compared with ANN-GA and ANN-PSO, respectively.

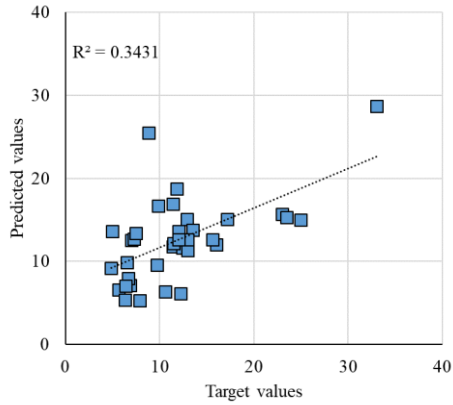 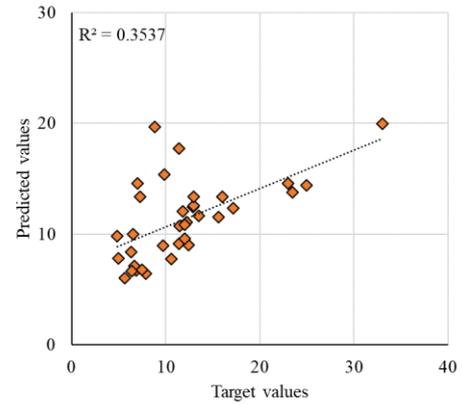

ANN-PSO　　　　　　　　　　　　　　　　ANN-GA

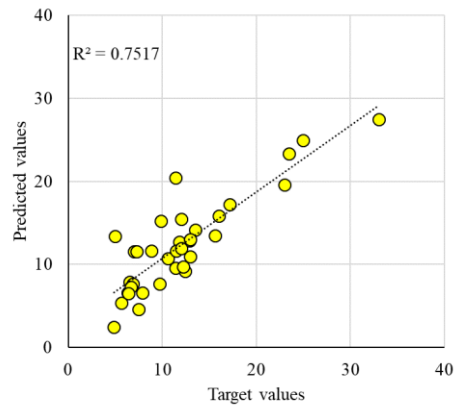

ANFIS-PSO

**Figure 8.** The plot diagrams of the developed models in testing phase

Figure 9 presents the deviation from target values for the predicted values from each model in the testing phase. This figure is developed for observing the differences of the target values from the predicted values by the models. According to Figure 9, the lowest deviation from target values is related to the ANN-PSO followed by ANN-GA. Therefore, it can be claimed that ANFIS-PSO can reduce the differences between predicted and target values, and accordingly increase the prediction performance, compared with ANN-GA and ANN-PSO.

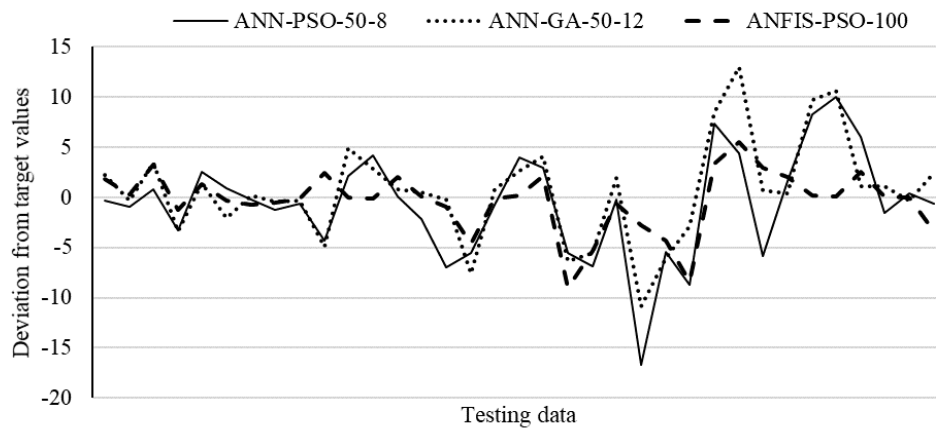

**Figure 9.** deviation from target values for the developed models

Figure 10 presents the Taylor diagram for the developed models. Taylor diagram compares the models in terms of correlation coefficient and standard deviation. Lower standard deviation and higher correlation coefficient refer to the best predicted results.

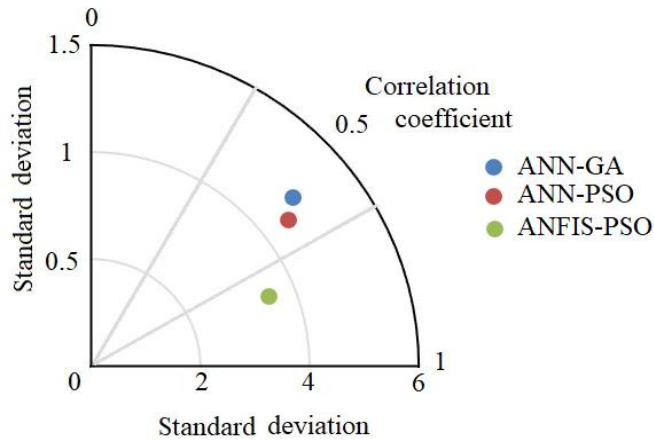

**Figure 10.** The Taylor diagram for the predicted values by the models

According to Figure 10, ANFIS-PSO can be selected as the best model for the prediction of the dependent value in comparison with ANN-GA and ANN-PSO.

## 4. Conclusions

In the present study, air demand in six dams has been investigated by physical scaled model in applied hydrodynamic laboratory of IUST. To predict air demand in downstream tunnels, more than 110 data series including the air velocity, water volume rate and gate opening percentage were collected. As conventional algorithms in term of single model have not predicted sufficient for this type of research, three robust hybrid ML methods were employed to predict the flow behavior in the bottom outlet of dams. The models include ANN and ANFIS which was trained using GA and PSO algorithms. The simulation results indicate that ANFIS-PSO in comparison with ANN-PSO and ANN-GA has the best error rate that RMSE for ANFIS-PSO, ANN-PSO and ANN-GA shows 3.097, 4.977 and 5.208, respectively. Taylor diagram compares the models in terms of correlation coefficient and standard deviation. Therefore, lower standard deviation and higher correlation coefficient refer to the best predicted results for ANFIS-PSO.

Nomenclature

| | | | |
|---|---|---|---|
| $Q_{water}$ | Water volume rate | $A_0$ | Maximum area of exit cross section of gate |
| $Q_{air}$ | Air volume rate | ANFIS | Adaptive neuro fuzzy inference system |
| β | Volume ratio | GA | Genetic algorithm |
| Fr | Froude number | PSO | Particle swarm optimization |
| μ | Water dynamic viscosity | MF | Membership function |
| H | Dam head | RMSE | Root mean square error |
| ϱ | Density of water | MSE | Mean square error |
| A | Exit cross-section of gate | SI | Sustainability index |
| $z_c$ | Height of minimum section | CC | Correlation coefficient |
| g | Gravitational acceleration | ANN | Artificial neural network |

**Acknowledgements**

We also deeply thank to Arvin Share Pardaz company to provide all experimental data and other information about six dams to carry out this research.